\documentclass[letterpaper,12pt,english]{article}
\usepackage[utf8]{inputenc}
\DeclareUnicodeCharacter{00A0}{\nobreakspace}
\usepackage{cmap}
\usepackage[T1]{fontenc}
\usepackage{babel}
\usepackage{times}
\usepackage[Bjarne]{fncychap}
\usepackage{longtable}
\usepackage{sphinx}
\usepackage{multirow}
\usepackage{eqparbox}
\usepackage{amsfonts}

\addto\captionsenglish{}
\addto\captionsenglish{}
\SetupFloatingEnvironment{literal-block}{name=Listing }

  \pdfoutput=1
  \usepackage{amssymb}
  \newcommand{\chapter}[1]{}  
  \newcommand{\ignore}[1]{}  

  \newcommand{\R}{\ensuremath{\mathbb{R}}}

  \newcommand{\IHV}{I_\mathrm{HV}^\mathrm{COCO}}

\title{Biobjective Performance Assessment with the COCO Platform}
\date{May 05, 2016}
\release{0.7-beta}
\date{\vspace{-1ex}}\author{
      Dimo Brockhoff$^1$,
      Tea Tu\v{s}ar$^1$, 
      Dejan Tu\v{s}ar$^1$, 
      Tobias Wagner$^2$,
      \\ Nikolaus Hansen$^{3,4}$,
      Anne Auger$^{3,4}$
  \\[1.5ex]
    $^1$Inria, research centre Lille, 
    $^2$Technical University Dortmund, 
    \\
    $^3$Inria, research centre Saclay, 
    $^4$Universit\'e Paris-Saclay, LRI 
    }

\renewcommand{\releasename}{Release}

\makeindex

\makeatletter
\def\PYG@reset{\let\PYG@it=\relax \let\PYG@bf=\relax%
    \let\PYG@ul=\relax \let\PYG@tc=\relax%
    \let\PYG@bc=\relax \let\PYG@ff=\relax}
\def\PYG@tok#1{\csname PYG@tok@#1\endcsname}
\def\PYG@toks#1+{\ifx\relax#1\empty\else%
    \PYG@tok{#1}\expandafter\PYG@toks\fi}
\def\PYG@do#1{\PYG@bc{\PYG@tc{\PYG@ul{%
    \PYG@it{\PYG@bf{\PYG@ff{#1}}}}}}}
\def\PYG#1#2{\PYG@reset\PYG@toks#1+\relax+\PYG@do{#2}}

\expandafter\def\csname PYG@tok@gd\endcsname{\def\PYG@tc##1{\textcolor[rgb]{0.63,0.00,0.00}{##1}}}
\expandafter\def\csname PYG@tok@gu\endcsname{\let\PYG@bf=\textbf\def\PYG@tc##1{\textcolor[rgb]{0.50,0.00,0.50}{##1}}}
\expandafter\def\csname PYG@tok@gt\endcsname{\def\PYG@tc##1{\textcolor[rgb]{0.00,0.27,0.87}{##1}}}
\expandafter\def\csname PYG@tok@gs\endcsname{\let\PYG@bf=\textbf}
\expandafter\def\csname PYG@tok@gr\endcsname{\def\PYG@tc##1{\textcolor[rgb]{1.00,0.00,0.00}{##1}}}
\expandafter\def\csname PYG@tok@cm\endcsname{\let\PYG@it=\textit\def\PYG@tc##1{\textcolor[rgb]{0.25,0.50,0.56}{##1}}}
\expandafter\def\csname PYG@tok@vg\endcsname{\def\PYG@tc##1{\textcolor[rgb]{0.73,0.38,0.84}{##1}}}
\expandafter\def\csname PYG@tok@vi\endcsname{\def\PYG@tc##1{\textcolor[rgb]{0.73,0.38,0.84}{##1}}}
\expandafter\def\csname PYG@tok@mh\endcsname{\def\PYG@tc##1{\textcolor[rgb]{0.13,0.50,0.31}{##1}}}
\expandafter\def\csname PYG@tok@cs\endcsname{\def\PYG@tc##1{\textcolor[rgb]{0.25,0.50,0.56}{##1}}\def\PYG@bc##1{\setlength{\fboxsep}{0pt}\colorbox[rgb]{1.00,0.94,0.94}{\strut ##1}}}
\expandafter\def\csname PYG@tok@ge\endcsname{\let\PYG@it=\textit}
\expandafter\def\csname PYG@tok@vc\endcsname{\def\PYG@tc##1{\textcolor[rgb]{0.73,0.38,0.84}{##1}}}
\expandafter\def\csname PYG@tok@il\endcsname{\def\PYG@tc##1{\textcolor[rgb]{0.13,0.50,0.31}{##1}}}
\expandafter\def\csname PYG@tok@go\endcsname{\def\PYG@tc##1{\textcolor[rgb]{0.20,0.20,0.20}{##1}}}
\expandafter\def\csname PYG@tok@cp\endcsname{\def\PYG@tc##1{\textcolor[rgb]{0.00,0.44,0.13}{##1}}}
\expandafter\def\csname PYG@tok@gi\endcsname{\def\PYG@tc##1{\textcolor[rgb]{0.00,0.63,0.00}{##1}}}
\expandafter\def\csname PYG@tok@gh\endcsname{\let\PYG@bf=\textbf\def\PYG@tc##1{\textcolor[rgb]{0.00,0.00,0.50}{##1}}}
\expandafter\def\csname PYG@tok@ni\endcsname{\let\PYG@bf=\textbf\def\PYG@tc##1{\textcolor[rgb]{0.84,0.33,0.22}{##1}}}
\expandafter\def\csname PYG@tok@nl\endcsname{\let\PYG@bf=\textbf\def\PYG@tc##1{\textcolor[rgb]{0.00,0.13,0.44}{##1}}}
\expandafter\def\csname PYG@tok@nn\endcsname{\let\PYG@bf=\textbf\def\PYG@tc##1{\textcolor[rgb]{0.05,0.52,0.71}{##1}}}
\expandafter\def\csname PYG@tok@no\endcsname{\def\PYG@tc##1{\textcolor[rgb]{0.38,0.68,0.84}{##1}}}
\expandafter\def\csname PYG@tok@na\endcsname{\def\PYG@tc##1{\textcolor[rgb]{0.25,0.44,0.63}{##1}}}
\expandafter\def\csname PYG@tok@nb\endcsname{\def\PYG@tc##1{\textcolor[rgb]{0.00,0.44,0.13}{##1}}}
\expandafter\def\csname PYG@tok@nc\endcsname{\let\PYG@bf=\textbf\def\PYG@tc##1{\textcolor[rgb]{0.05,0.52,0.71}{##1}}}
\expandafter\def\csname PYG@tok@nd\endcsname{\let\PYG@bf=\textbf\def\PYG@tc##1{\textcolor[rgb]{0.33,0.33,0.33}{##1}}}
\expandafter\def\csname PYG@tok@ne\endcsname{\def\PYG@tc##1{\textcolor[rgb]{0.00,0.44,0.13}{##1}}}
\expandafter\def\csname PYG@tok@nf\endcsname{\def\PYG@tc##1{\textcolor[rgb]{0.02,0.16,0.49}{##1}}}
\expandafter\def\csname PYG@tok@si\endcsname{\let\PYG@it=\textit\def\PYG@tc##1{\textcolor[rgb]{0.44,0.63,0.82}{##1}}}
\expandafter\def\csname PYG@tok@s2\endcsname{\def\PYG@tc##1{\textcolor[rgb]{0.25,0.44,0.63}{##1}}}
\expandafter\def\csname PYG@tok@nt\endcsname{\let\PYG@bf=\textbf\def\PYG@tc##1{\textcolor[rgb]{0.02,0.16,0.45}{##1}}}
\expandafter\def\csname PYG@tok@nv\endcsname{\def\PYG@tc##1{\textcolor[rgb]{0.73,0.38,0.84}{##1}}}
\expandafter\def\csname PYG@tok@s1\endcsname{\def\PYG@tc##1{\textcolor[rgb]{0.25,0.44,0.63}{##1}}}
\expandafter\def\csname PYG@tok@ch\endcsname{\let\PYG@it=\textit\def\PYG@tc##1{\textcolor[rgb]{0.25,0.50,0.56}{##1}}}
\expandafter\def\csname PYG@tok@m\endcsname{\def\PYG@tc##1{\textcolor[rgb]{0.13,0.50,0.31}{##1}}}
\expandafter\def\csname PYG@tok@gp\endcsname{\let\PYG@bf=\textbf\def\PYG@tc##1{\textcolor[rgb]{0.78,0.36,0.04}{##1}}}
\expandafter\def\csname PYG@tok@sh\endcsname{\def\PYG@tc##1{\textcolor[rgb]{0.25,0.44,0.63}{##1}}}
\expandafter\def\csname PYG@tok@ow\endcsname{\let\PYG@bf=\textbf\def\PYG@tc##1{\textcolor[rgb]{0.00,0.44,0.13}{##1}}}
\expandafter\def\csname PYG@tok@sx\endcsname{\def\PYG@tc##1{\textcolor[rgb]{0.78,0.36,0.04}{##1}}}
\expandafter\def\csname PYG@tok@bp\endcsname{\def\PYG@tc##1{\textcolor[rgb]{0.00,0.44,0.13}{##1}}}
\expandafter\def\csname PYG@tok@c1\endcsname{\let\PYG@it=\textit\def\PYG@tc##1{\textcolor[rgb]{0.25,0.50,0.56}{##1}}}
\expandafter\def\csname PYG@tok@o\endcsname{\def\PYG@tc##1{\textcolor[rgb]{0.40,0.40,0.40}{##1}}}
\expandafter\def\csname PYG@tok@kc\endcsname{\let\PYG@bf=\textbf\def\PYG@tc##1{\textcolor[rgb]{0.00,0.44,0.13}{##1}}}
\expandafter\def\csname PYG@tok@c\endcsname{\let\PYG@it=\textit\def\PYG@tc##1{\textcolor[rgb]{0.25,0.50,0.56}{##1}}}
\expandafter\def\csname PYG@tok@mf\endcsname{\def\PYG@tc##1{\textcolor[rgb]{0.13,0.50,0.31}{##1}}}
\expandafter\def\csname PYG@tok@err\endcsname{\def\PYG@bc##1{\setlength{\fboxsep}{0pt}\fcolorbox[rgb]{1.00,0.00,0.00}{1,1,1}{\strut ##1}}}
\expandafter\def\csname PYG@tok@mb\endcsname{\def\PYG@tc##1{\textcolor[rgb]{0.13,0.50,0.31}{##1}}}
\expandafter\def\csname PYG@tok@ss\endcsname{\def\PYG@tc##1{\textcolor[rgb]{0.32,0.47,0.09}{##1}}}
\expandafter\def\csname PYG@tok@sr\endcsname{\def\PYG@tc##1{\textcolor[rgb]{0.14,0.33,0.53}{##1}}}
\expandafter\def\csname PYG@tok@mo\endcsname{\def\PYG@tc##1{\textcolor[rgb]{0.13,0.50,0.31}{##1}}}
\expandafter\def\csname PYG@tok@kd\endcsname{\let\PYG@bf=\textbf\def\PYG@tc##1{\textcolor[rgb]{0.00,0.44,0.13}{##1}}}
\expandafter\def\csname PYG@tok@mi\endcsname{\def\PYG@tc##1{\textcolor[rgb]{0.13,0.50,0.31}{##1}}}
\expandafter\def\csname PYG@tok@kn\endcsname{\let\PYG@bf=\textbf\def\PYG@tc##1{\textcolor[rgb]{0.00,0.44,0.13}{##1}}}
\expandafter\def\csname PYG@tok@cpf\endcsname{\let\PYG@it=\textit\def\PYG@tc##1{\textcolor[rgb]{0.25,0.50,0.56}{##1}}}
\expandafter\def\csname PYG@tok@kr\endcsname{\let\PYG@bf=\textbf\def\PYG@tc##1{\textcolor[rgb]{0.00,0.44,0.13}{##1}}}
\expandafter\def\csname PYG@tok@s\endcsname{\def\PYG@tc##1{\textcolor[rgb]{0.25,0.44,0.63}{##1}}}
\expandafter\def\csname PYG@tok@kp\endcsname{\def\PYG@tc##1{\textcolor[rgb]{0.00,0.44,0.13}{##1}}}
\expandafter\def\csname PYG@tok@w\endcsname{\def\PYG@tc##1{\textcolor[rgb]{0.73,0.73,0.73}{##1}}}
\expandafter\def\csname PYG@tok@kt\endcsname{\def\PYG@tc##1{\textcolor[rgb]{0.56,0.13,0.00}{##1}}}
\expandafter\def\csname PYG@tok@sc\endcsname{\def\PYG@tc##1{\textcolor[rgb]{0.25,0.44,0.63}{##1}}}
\expandafter\def\csname PYG@tok@sb\endcsname{\def\PYG@tc##1{\textcolor[rgb]{0.25,0.44,0.63}{##1}}}
\expandafter\def\csname PYG@tok@k\endcsname{\let\PYG@bf=\textbf\def\PYG@tc##1{\textcolor[rgb]{0.00,0.44,0.13}{##1}}}
\expandafter\def\csname PYG@tok@se\endcsname{\let\PYG@bf=\textbf\def\PYG@tc##1{\textcolor[rgb]{0.25,0.44,0.63}{##1}}}
\expandafter\def\csname PYG@tok@sd\endcsname{\let\PYG@it=\textit\def\PYG@tc##1{\textcolor[rgb]{0.25,0.44,0.63}{##1}}}

\makeatother

\begin{document}

\maketitle
\phantomsection\label{index::doc}

\chapter{CHAPTERTITLE}
\label{index:biobjective-performance-assessment-with-the-coco-platform}\label{index:chaptertitle}
\begin{abstract}
This document details the rationales behind assessing the performance of
numerical black-box optimizers on multi-objective problems within the \href{https://github.com/numbbo/coco}{COCO}
platform and in particular on the biobjective test suite \href{http://numbbo.github.io/coco-doc/bbob-biobj/functions}{\code{bbob-biobj}}.
The evaluation is based on a hypervolume of all non-dominated solutions in the
archive of candidate solutions and measures the runtime until the
hypervolume value succeeds prescribed target values.
\end{abstract}\tableofcontents 
\newpage

\section{Introduction}
\label{index:introduction}
The performance assessment of (numerical) optimization algorithms with the \href{https://github.com/numbbo/coco}{COCO}
platform \phantomsection\label{index:id1}{\hyperref[index:han2016co]{\emph{{[}HAN2016co{]}}}} is invariably based on the
measurement of the \emph{runtime}\footnote[1]{
Time is considered to be \emph{number of function evaluations} and,
consequently, runtime is measured in number of function evaluations.
} until a \emph{quality indicator} reaches a predefined
\emph{target value}.
On each problem instance, several target values are defined and for each
target value a runtime is measured (or no runtime value is available if the
indicator does not reach the target value) \phantomsection\label{index:id3}{\hyperref[index:han2016perf]{\emph{{[}HAN2016perf{]}}}}.
In the single-objective, noise-free case, the assessed quality indicator is, at
each given time step, the function value of the best solution the algorithm has
obtained (evaluated or recommended, see \phantomsection\label{index:id4}{\hyperref[index:han2016ex]{\emph{{[}HAN2016ex{]}}}}) before or at this time
step.

In the bi- and multi-objective case, e.g. on the biobjective \code{bbob-biobj}
test suite \phantomsection\label{index:id5}{\hyperref[index:tus2016]{\emph{{[}TUS2016{]}}}}, the assessed quality
indicator at the given time step is a hypervolume indicator computed from
\emph{all} solutions obtained (evaluated or recommended) before or at this time
step.

\subsection{Definitions and Terminology}
\label{index:definitions-and-terminology}
In this section, we introduce the definitions of some basic terms and concepts.
\begin{description}
\item[{\emph{function instance, problem}}] \leavevmode
In the case of the bi-objective performance assessment within \href{https://github.com/numbbo/coco}{COCO}, a problem is a 5-tuple of
\begin{itemize}
\item {} 
a \emph{parameterized function} \(f_\theta: \mathbb{R}^n \to \mathbb{R}^2\), mapping the decision variables of a solution \(x\in\mathbb{R}^n\) to its objective vector \(f_\theta(x) = (f_\alpha(x),f_\beta(x))\) with \(f_\alpha: \mathbb{R}^n \mapsto \mathbb{R}\) and \(f_\beta: \mathbb{R}^n \mapsto \mathbb{R}\) being parameterized (single-objective) functions themselves

\item {} 
its concrete parameter value \(\theta\in\Theta\) determining the so-called
\emph{function instance} \(i\),

\item {} 
the \emph{problem dimension} \(n\),

\item {} 
an underlying quality indicator \(I\), mapping a set of solutions to its quality, and

\item {} 
a \emph{target value} \(I_{\rm target}\) of the underlying quality indicator, see below for details.

\end{itemize}

We call a problem \emph{solved} by an optimization algorithm if the algorithm
reaches a quality indicator value at least as good as the associated target value.
The number of function evaluations needed to surpass the target value for the first time
is \href{https://github.com/numbbo/coco}{COCO}`s central performance measure. \phantomsection\label{index:id7}{\hyperref[index:han2016co]{\emph{{[}HAN2016co{]}}}} In case a single
quality indicator is used for all problems in a benchmark suite, we can drop the
quality indicator and refer to a problem as a quadruple \(f_\theta,\theta,n,I_{\rm target}\).
Note that typically more than one problem for a \emph{function instance} of
\((f_\theta,\theta,n)\) is defined by choosing more than one target value.

\item[{\emph{Pareto set}, \emph{Pareto front}, and \emph{Pareto dominance}}] \leavevmode
For a function instance, i.e., a function \(f_\theta=(f_\alpha,f_\beta)\) with
given parameter value \(\theta\) and dimension \(n\), the Pareto set is the set
of all (Pareto-optimal) solutions for which no solutions in the search space
\(\R^n\) exist that have either an improved \(f_\alpha\) or an improved
\(f_\beta\) value while the other value is at least as good
(or in other words, a \emph{Pareto-optimal} solution in the Pareto set has no other solution
that \emph{dominates} it). The image of the Pareto set in the \emph{objective space} is called
the Pareto front. We generalize the standard Pareto dominance relation to sets by saying
solution set \(A=\{a_1,\ldots,a_{|A|}\}\) dominates solution set \(B=\{b_1,\ldots,b_{|B|}\}\)
if and only if for all \(b_i\in B\) there is at least one solution \(a_j\)
that dominates it.

\item[{\emph{ideal point}}] \leavevmode
The ideal point (in objective space) is defined as the vector in objective space that
contains the optimal function value for each objective \emph{independently}, i.e. for the above
concrete function instance, the ideal point is given by
\(z_{\rm ideal}  = (\inf_{x\in \mathbb{R}^n} f_\alpha(x), \inf_{x\in \mathbb{R}^n} f_\beta(x))\).

\item[{\emph{nadir point}}] \leavevmode
The nadir point (in objective space) consists in each objective of
the worst value obtained by any Pareto-optimal solution. More precisely, if
\(\mathcal{PO}\) denotes the Pareto set, the nadir point satisfies
\(z_{\rm nadir}  =  \left( \sup_{x \in \mathcal{PO}} f_\alpha(x),
\sup_{x \in \mathcal{PO}} f_\beta(x)  \right)\).

\item[{\emph{archive}}] \leavevmode
An external archive or simply an archive is the set of non-dominated solutions,
obtained over an algorithm run. At each point \(t\) in time (that is after
\(t\) function evaluations), we consider the set of all
mutually non-dominating solutions that have been evaluated so far. We
denote the archive after \(t\) function evaluations as \(A_t\)
and use it to define the performance of the algorithm in terms of a (quality)
indicator function \(A_t \rightarrow \R\) that might depend on a problem's
underlying parameterized function and its dimension and instance.

\end{description}

\section{Performance Assessment with a Quality Indicator}
\label{index:performance-assessment-with-a-quality-indicator}
For measuring the runtime on a given problem, we consider a quality indicator
which is to be optimized (minimized).
In the noiseless single-objective case, the quality indicator is the best so-far observed objective function value (recommendations can replace previous observations).
In the case of the \code{bbob-biobj} test suite, the quality indicator is based on the
hypervolume indicator of the \emph{archive} \(A_t\).

\subsection{Definition of the Quality Indicator}
\label{index:definition-of-the-quality-indicator}
The indicator \(\IHV\) to be mininized is either the negative
hypervolume indicator of the archive with the nadir
point as reference point or the distance to the region of interest
\([z_{\text{ideal}}, z_{\text{nadir}}]\) after a normalization of the
objective space\footnote[2]{
We conduct an affine transformation of both objective function values
such that the ideal point \(z_{\text{ideal}}= (z_{\text{ideal}, \alpha},
z_{\text{ideal}, \beta})\) is mapped to \((0,0)\) and the nadir point
\(z_{\text{nadir}}= (z_{\text{nadir}, \alpha}, z_{\text{nadir}, \beta})\)
is mapped to \((1,1)\).
}:
    \begin{equation*}
    \IHV =  \left\{ \begin{array}{ll}
    - \text{HV}(A_t, [z_{\text{ideal}}, z_{\text{nadir}}]) & \text{if $A_t$ dominates } \{z_{\text{nadir}}\}\\
    dist(A_t, [z_{\text{ideal}}, z_{\text{nadir}}]) & \text{otherwise}
    \end{array}     \right.\enspace .
    \end{equation*}
where
\begin{equation*}
\text{HV}(A_t, z_{\text{ideal}}, z_{\text{nadir}}) = \text{VOL}\left( \bigcup_{a \in A_t} \left[\frac{f_\alpha(a)-z_{\text{ideal}, \alpha}}{z_{\text{nadir}, \alpha}-z_{\text{ideal}, \alpha}}, 1\right]\times\left[\frac{f_\beta(a)-z_{\text{ideal}, \beta}}{z_{\text{nadir}, \beta}-z_{\text{ideal}, \beta}}, 1\right]\right)
    \end{equation*}
is the (normalized) hypervolume of archive \(A_t\) with respect to the
nadir point \((z_{\text{nadir}, \alpha}, z_{\text{nadir},\beta})\) as reference point and where (with division understood to be element-wise, Hadamard division),
\begin{equation*}
    dist(A_t, [z_{\text{ideal}}, z_{\text{nadir}}]) = \inf_{a\in A_t, z\in [z_{\text{ideal}}, z_{\text{nadir}}]} \left\|\frac{f(a)-z}{z_{\text{nadir}}-z_{\text{ideal}}}\right\|
    \end{equation*}
is the smallest (normalized) Euclidean distance between a solution in the archive and the region of interest, see also the figures below for an illustration.
\begin{figure}[htbp]
\centering
\capstart

\includegraphics[width=0.600\linewidth]{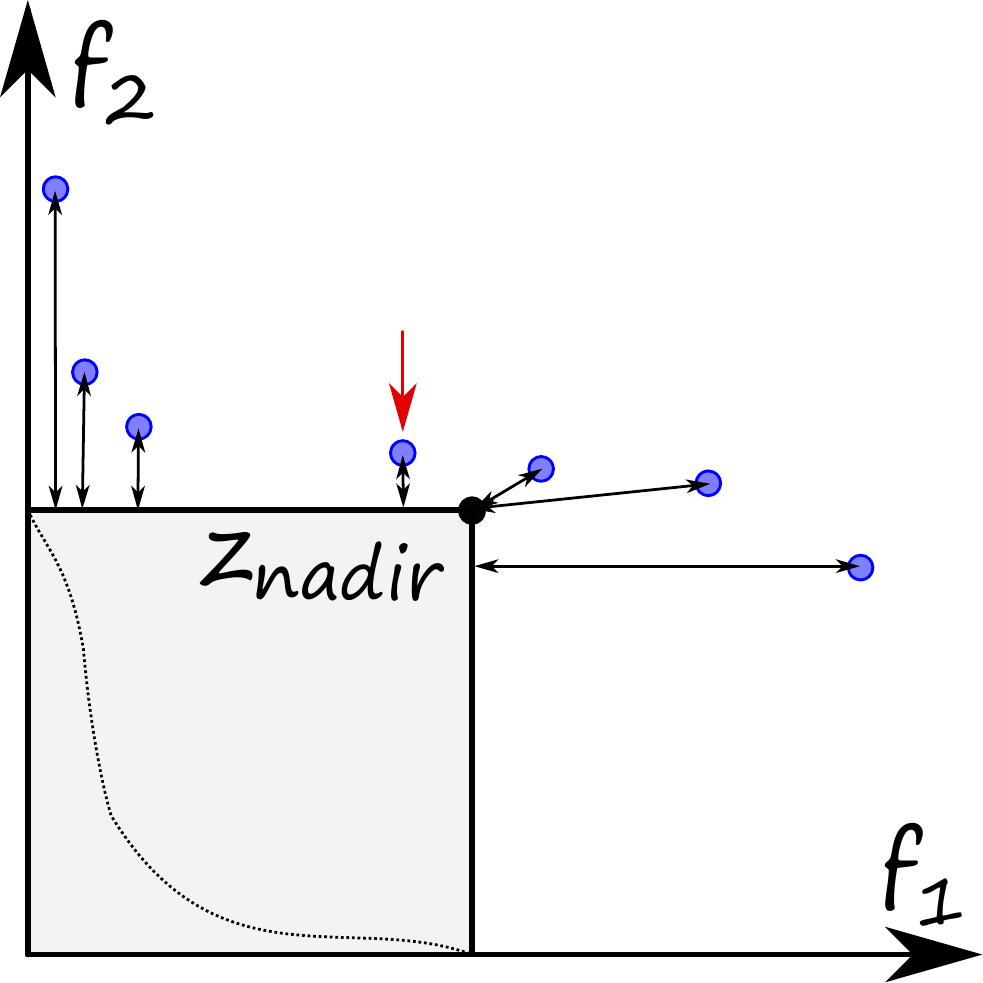}
\caption{Illustration of Coco's quality indicator (to be minimized) in the
(normalized) bi-objective case if no solution of the archive (blue filled circles)
dominates the nadir point (black filled circle), i.e., the shortest
distance of an archive member to the region of interest (ROI), delimited
by the nadir point.
Here, it is the fourth point from the left (indicated by the red arrow) that defines
the smallest distance.}\end{figure}
\begin{figure}[htbp]
\centering
\capstart

\includegraphics[width=0.600\linewidth]{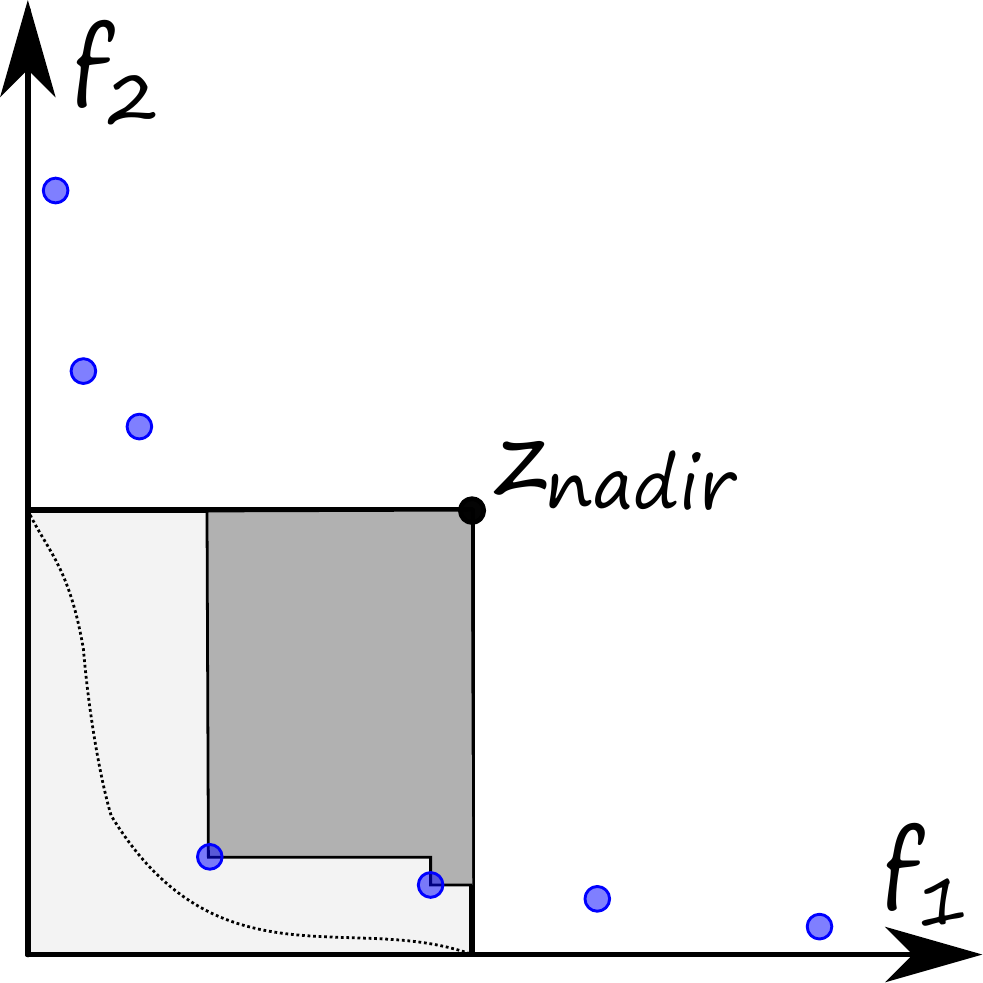}
\caption{Illustration of Coco's quality indicator (to be minimized) in the
bi-objective case if the nadir point (black filled circle) is dominated by
at least one solution in the archive (blue filled circles). The indicator is the
negative hypervolume of the archive with the nadir point as reference point.}\end{figure}

\subsection{Rationales Behind the Performance Measure}
\label{index:rationales-behind-the-performance-measure}\begin{description}
\item[{\emph{Why using an archive?}}] \leavevmode
We believe using an archive to keep all non-dominated solutions is relevant in practice
in bi-objective real-world applications, in particular when function evaluations are
expensive. Using an external archive for the performance assessment has the additional
advantage that no populuation size needs to be prescribed and algorithms with different
or even changing population sizes can be easily compared.

\item[{\emph{Why hypervolume?}}] \leavevmode
Although, in principle, other quality indicators can be used in replacement of the
hypervolume, the monotonicity of the hypervolume is a strong theoretical argument
for using it in the performance assessment: the hypervolume indicator value of the
archive improves if and only if a new non-dominated solution is generated \phantomsection\label{index:id10}{\hyperref[index:zit2003]{\emph{{[}ZIT2003{]}}}}.

\end{description}

\subsection{Specificities and Properties}
\label{index:specificities-and-properties}
In summary, the proposed \code{bbob-biobj} performance criterion has the following
specificities:
\begin{itemize}
\item {} 
Algorithm performance is measured via runtime until the quality of the archive of non-dominated
solutions found so far surpasses a target value.

\item {} 
To compute the quality indicator, the objective space is normalized.
The region of interest (ROI) \([z_{\text{ideal}}, z_{\text{nadir}}]\),
defined by the ideal and nadir point, is mapped to \([0, 1]^2\).

\item {} 
If the nadir point is dominated by at least one point in the archive, the
quality is computed as the negative hypervolume of the archive using
the nadir point as hypervolume reference point.

\item {} 
If the nadir point is not dominated by the archive, the quality equals the
distance of the archive to the ROI.

\end{itemize}

This implies that:
\begin{itemize}
\item {} 
the quality indicator value of an archive that contains the nadir point as
non-dominated point is \(0\).

\item {} 
the quality indicator value is bounded from below by \(-1\), which is
the quality of an archive that contains the ideal point, and

\item {} 
because the quality of an archive is used as performance criterion, no
population size has to be prescribed to the algorithm. In particular,
steady-state and generational algorithms can be compared directly as well
as algorithms with varying population size and algorithms which carry along
their external archive themselves.

\end{itemize}

\section{Definition of Target Values}
\label{index:definition-of-target-values}
For each problem instance of the benchmark suite, consisting of a parameterized
function, its dimension and its instance parameter \(\theta_i\), a set of quality
indicator target values is chosen, eventually used to measure algorithm runtime to
reach each of these targets.
The target values are based on a target precision \(\Delta I\) and a
\emph{reference hypervolume indicator value}, \(I_i^\mathrm{ref}\), which is an approximation of the
\(\IHV\) indicator value of the Pareto set.

\subsection{Target Precision Values}
\label{index:target-precision-values}
All target indicator values are computed in the form of \(I_i^\mathrm{ref}\) \(+\,\Delta
I\) from the instance dependent reference value \(I_i^\mathrm{ref}\) and a target precision
value \(\Delta I\).
For the \code{bbob-biobj} test suite, 58 target precisions \(\Delta I\) are
chosen, identical for all problem instances, as
\begin{gather}
\begin{split}\Delta I \in \{ \underbrace{-10^{-4}, -10^{-4.2}, \dots, -10^{-4.8}, -10^{-5}}_{
\text{six negative target precision values}}, 0, 10^{-5}, 10^{-4.9}, 10^{-4.8}, \dots, 10^{-0.1}, 10^0 \}\enspace.\end{split}\notag
\end{gather}
Negative target precisions are used because the reference indicator value, as
defined in the next section, can be surpassed by an optimization algorithm.\footnote[3]{
In comparison, the reference value in the single-objective case is
the \(f\)-value of the known global optimum and, consequently, the target
precision values have been strictly positive \phantomsection\label{index:id14}{\hyperref[index:han2016perf]{\emph{{[}HAN2016perf{]}}}}.
}
The runtimes to reach these target values are presented as empirical cumulative
distribution function, ECDF \phantomsection\label{index:id12}{\hyperref[index:han2016perf]{\emph{{[}HAN2016perf{]}}}}.
Runtimes to reach specific target precisions are presented as well.
It is not uncommon however that the quality indicator value of the algorithm
never surpasses some of these target values, which leads to missing runtime
measurements.

\subsection{The Reference Hypervolume Indicator Value}
\label{index:the-reference-hypervolume-indicator-value}
Unlike the single-objective \code{bbob} test suite \phantomsection\label{index:id15}{\hyperref[index:han2009fun]{\emph{{[}HAN2009fun{]}}}}, the
biobjective \code{bbob-biobj} test suite does not provide analytic expressions of
its optima.
Except for \(f_1\), the Pareto set and the Pareto front are unknown.

Instead of the unknown hypervolume of the true Pareto set, we use the hypervolume of an approximation of the Pareto set as reference hypervolume indicator value \(I_i^\mathrm{ref}\).\footnote[4]{
Using the quality indicator value of the \emph{true} Pareto set might not
be desirable, because the set contains an infinite number of solutions,
which is neither a possible nor a desirable goal to aspire in practice.
}
To obtain the approximation, several multi-objective optimization algorithms
have been run and all non-dominated solutions over all runs have been
recorded.\footnote[5]{
Amongst others, we run versions of NSGA-II \phantomsection\label{index:id20}{\hyperref[index:deb2002]{\emph{{[}DEB2002{]}}}} via Matlab's
\code{gamultiobj} \href{http://www.mathworks.com/help/gads/gamultiobj.html}{function}, SMS-EMOA \phantomsection\label{index:id21}{\hyperref[index:beu2007]{\emph{{[}BEU2007{]}}}}, MOEA/D \phantomsection\label{index:id22}{\hyperref[index:zha2007]{\emph{{[}ZHA2007{]}}}},
RM-MEDA \phantomsection\label{index:id23}{\hyperref[index:zha2008]{\emph{{[}ZHA2008{]}}}}, and MO-CMA-ES \phantomsection\label{index:id24}{\hyperref[index:vos2010]{\emph{{[}VOS2010{]}}}}, together with simple
uniform RANDOMSEARCH and the single-objective CMA-ES \phantomsection\label{index:id25}{\hyperref[index:han2001]{\emph{{[}HAN2001{]}}}} on scalarized problems
(i.e. weighted sum) to create first approximations of the bi-objective
problems' Pareto sets.
}
The hypervolume indicator value of the obtained set of non-dominated
solutions, also called \emph{non-dominated reference set}, separately obtained
for each problem instance in the benchmark suite, is then used as the
reference hypervolume indicator value.

\section{Instances and Generalization Experiment}
\label{index:instances-and-generalization-experiment}
The standard procedure for an experiment on a benchmark suite, like the
\code{bbob-biobj} suite, prescribes to run the algorithm of choice once on each
problem of the suite \phantomsection\label{index:id27}{\hyperref[index:han2016ex]{\emph{{[}HAN2016ex{]}}}}.
For the \code{bbob-biobj} suite, the postprocessing part of \href{https://github.com/numbbo/coco}{COCO} displays currently by
default only 5 out of the 10 instances from each function-dimension pair.

\section{Data Storage and Future Recalculations of Indicator Values}
\label{index:data-storage-and-future-recalculations-of-indicator-values}
Having a good approximation of the Pareto set/Pareto front is crucial in assessing
algorithm performance with the above suggested performance criterion. In order to allow
the reference sets to approximate the Pareto set/Pareto front better and better over time,
the \href{https://github.com/numbbo/coco}{COCO} platform records every non-dominated solution over the algorithm run.
Algorithm data sets, submitted through the \href{https://github.com/numbbo/coco}{COCO} platform's web page, can therefore
be used to improve the quality of the reference set by adding all solutions to the
reference set which are currently non-dominated to it.

Recording every new non-dominated solution within every algorithm run also allows to
recover the algorithm runs after the experiment and to recalculate the corresponding
hypervolume difference values if the reference set changes in the future. In order
to be able to distinguish between different collections of reference sets that might
have been used during the actual benchmarking experiment and the production of the
graphical output, \href{https://github.com/numbbo/coco}{COCO} writes the absolute hypervolume reference values together
with the performance data during the benchmarking experiment and displays
a version number in the plots generated that allows to retrieve the used reference
values from the \href{https://github.com/numbbo/coco}{Github repository of COCO}.
\section*{Acknowledgements}
This work was supported by the grant ANR-12-MONU-0009 (NumBBO)
of the French National Research Agency.

The authors would like to thank Thanh-Do Tran for his
contributions and assistance with the preliminary code of the bi-objective
setting and for providing us with his extensive experimental data. We also thank
Tobias Glasmachers, Oswin Krause, and Ilya Loshchilov for their bug reports, feature
requests, code testing, and many valuable discussions. Special thanks go
to Olaf Mersmann for the inital rewriting of the COCO platform without which
the bi-objective extension of COCO would not have happened.

\renewcommand{\indexname}{Index}
\printindex
\end{document}